\documentclass[11pt]{article}
\usepackage{float}
\usepackage[final]{acl}

\usepackage{times}
\usepackage{latexsym}

\usepackage[T1]{fontenc}

\usepackage[utf8]{inputenc}

\usepackage{microtype}

\usepackage{times}
\usepackage{latexsym}
\usepackage[T1]{fontenc}
\usepackage[utf8]{inputenc}
\usepackage{microtype}
\usepackage{inconsolata}
\usepackage{graphicx}
\usepackage{booktabs}
\usepackage{amsmath}
\usepackage{amssymb}
\usepackage[table]{xcolor}
\usepackage{colortbl}
\usepackage{multirow}
\usepackage{enumitem}
\definecolor{poscol}{HTML}{0E7C2D}
\definecolor{negcol}{HTML}{B71C1C}
\definecolor{posbg}{HTML}{E6F4EA}
\definecolor{negbg}{HTML}{FCE8E6}
\usepackage{algorithm}
\usepackage{algpseudocode}
\usepackage{float}
\usepackage{xurl}
\usepackage{array}

\usepackage{xcolor}
\usepackage{xspace}
\usepackage[capitalise]{cleveref}
\makeatletter
\AddToHook{cmd/appendix/before}{\def\cref@section@alias{appendix}}
\makeatother

\crefname{appendix}{Appendix}{Appendices}
\Crefname{appendix}{Appendix}{Appendices}
\crefname{subappendix}{Appendix}{Appendices}
\Crefname{subappendix}{Appendix}{Appendices}
\crefname{subsubappendix}{Appendix}{Appendices}
\Crefname{subsubappendix}{Appendix}{Appendices}

\let\oldappendix\appendix
\renewcommand{\appendix}{%
  \oldappendix%
  \crefalias{section}{appendix}%
  \crefalias{subsection}{subappendix}%
  \crefalias{subsubsection}{subsubappendix}%
}

\usepackage{listings}
\lstdefinestyle{prompt}{
  basicstyle=\footnotesize\ttfamily,
  breaklines=true,
  breakindent=0pt,
  columns=fullflexible,
  keepspaces=true,
  frame=single,
  framesep=5pt,
  xleftmargin=4pt,
  xrightmargin=4pt,
  aboveskip=6pt,
  belowskip=6pt,
  backgroundcolor=\color{black!3},
  rulecolor=\color{black!40},
}

\lstdefinestyle{trace}{
  basicstyle=\footnotesize\ttfamily,
  breaklines=true,
  breakindent=0pt,
  columns=fullflexible,
  keepspaces=true,
  frame=single,
  framesep=5pt,
  xleftmargin=4pt,
  xrightmargin=4pt,
  aboveskip=6pt,
  belowskip=6pt,
  backgroundcolor=\color{black!3},
  rulecolor=\color{black!40},
  showstringspaces=false,
  morestring=[b]",
  stringstyle=\color{violet!55!black},
  morecomment=[l][\color{black!45}\itshape]{\#},
  morekeywords={retrieve,refine,decompose,commit},
  keywordstyle=\color{blue!60!black}\bfseries,
  literate=*
    {[WRONG]}{{\color{negcol}\bfseries[WRONG]}}{7}
    {[CORRECT]}{{\color{poscol}\bfseries[CORRECT]}}{9}
    {-tel}{{\color{negcol}\bfseries-tel}}{4}
    {+tel}{{\color{poscol}\bfseries+tel}}{4},
}

\usepackage{inconsolata}
\newcommand{\method}[0]{\textsc{CalVerT}\xspace}
\newcommand{\methodlong}[0]{\underline{Cal}ibrated \underline{Ver}ifier \underline{T}elemetry}
\newcommand{\dinco}[0]{\textsc{DiNCo}\xspace}

\usepackage[colorinlistoftodos]{todonotes}

\newcommand{\myparagraph}[1]{\noindent \textbf{#1\hspace{0.3em}}}

\title{\method: Augmenting Agents with Calibrated Verifier Telemetry Improves Action and Learning in Knowledge-Intensive Tasks}

\author{Ashwin Vinod \qquad Ying Ding \qquad Elias Stengel-Eskin \\
The University of Texas at Austin}

\begin{document}
\maketitle
\begin{abstract}
LLM agents in knowledge intensive question answering take retrieval and reasoning actions with incomplete knowledge about whether their current answer is uncertain, unsupported, or already complete. 
This produces two failure modes: committing to confident but unsupported answers, which hurts accuracy, and over-retrieving when the evidence in hand already suffices, resulting in wasted compute. 
To give agents a more complete picture of the state space they are operating in, we introduce \textbf{calibrated verifier telemetry} (\method{}), which augments the agent's state with additional telemetry: a calibrated self-confidence score and a grounding verifier score. 
We show that \method{} can improve agents in both training-free and training-based settings. 
On four QA benchmarks, we find that \method{} raises F1 by triggering retrieval in cases where agents over-rely on parametric knowledge, while cutting redundant retrieval in cases where agents have sufficient context to answer. 
We show that \method{} can augment existing QA frameworks without training.
Moreover, \method{} also improves trained systems: by simply augmenting an agent's state with telemetry, we observe improvements after reinforcement learning, as compared to an agent with identical training but no \method{} telemetry.\footnote{Code: \href{https://github.com/ashwinn-v/CalVerT}{https://github.com/ashwinn-v/CalVerT}}
\end{abstract}

\section{Introduction}
Large language model (LLM) agents increasingly solve knowledge-intensive question answering (QA) and tool-use tasks by repeatedly retrieving, reflecting, and revising \citep{yao2023react,trivedi2023ircot,asai2024selfrag,zhou2024lats}.
Knowing when to perform these actions is critical; however,
 many current agents fail according to two failure modes: 
\textbf{parametric over-trust}, where agents over-rely on their potentially faulty parametric knowledge, and \textbf{over-retrieval}, where agents continue retrieving passages after already having sufficient evidence. 
The former poses a risk to answer correctness, while the latter leads to higher latency and cost (though it can also hurt answer quality).
We argue that existing agents \citep{deng2023mind2web, zhou2024webarena, yang2024sweagent, qin2024toolllm} are often fighting an uphill battle in balancing these concerns: they generally lack the \emph{measurement capabilities} to know whether the current answer is supported by the evidence in hand or whether the model itself is confident in it.

\begin{figure}[t]
\centering
\includegraphics[width=\columnwidth]{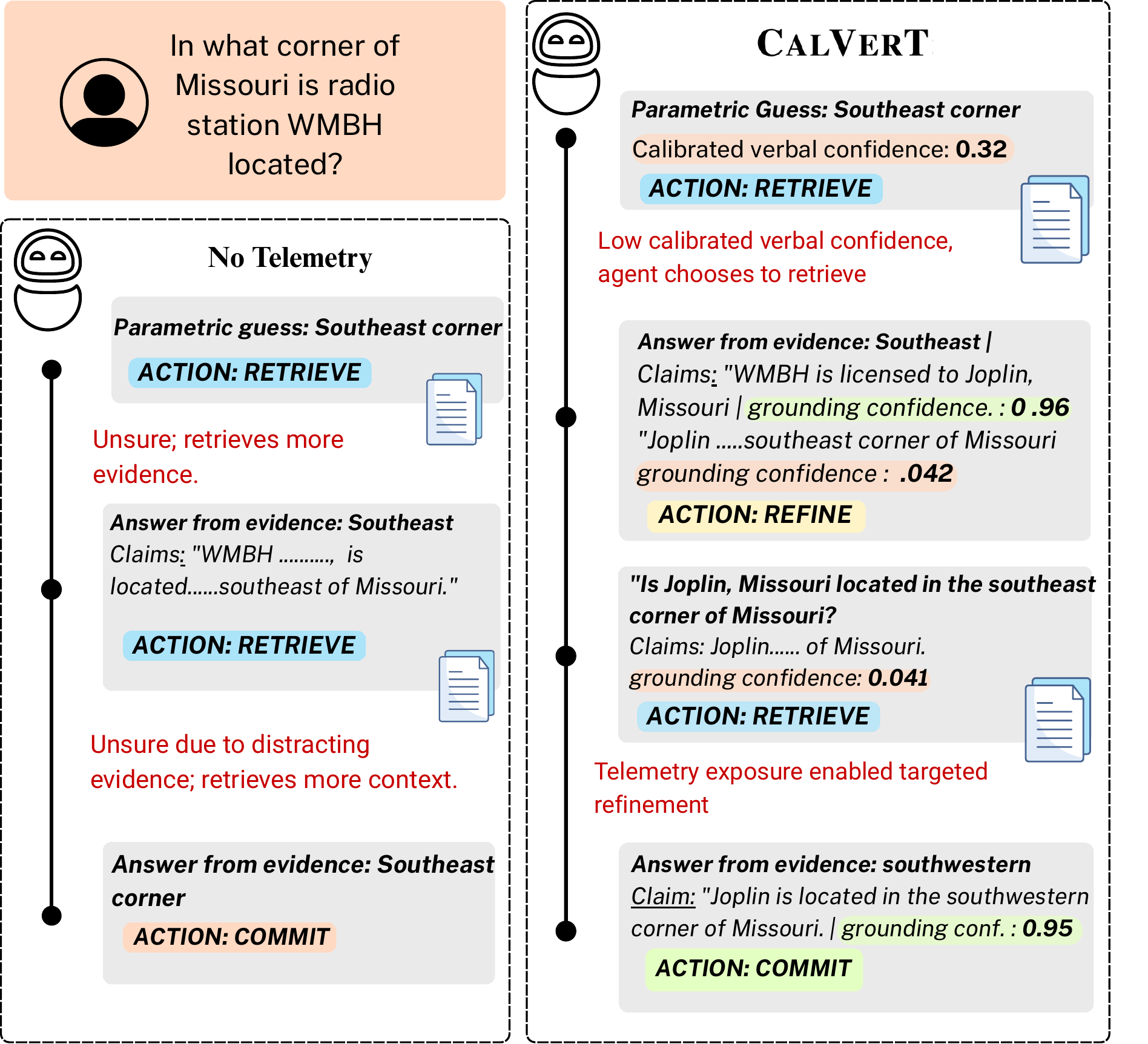}
\vspace{-1.5em}
\caption{Exposing calibrated confidence and grounding signals to the agent each turn yields better action choices. \textit{Left}: without telemetry, the agent retrieves additional evidence but still stops with an unsupported answer for the subquestion, causing an incorrect final answer. \textit{Right}: \method exposes low confidence and weak grounding in the form of verifier telemetry, so the agent first retrieves when the current evidence is insufficient, then refines the answer for targeted retrieval, and only commits after the answer is well grounded.
}
\vspace{-1.0em}
\label{fig:teaser}
\end{figure}

To address this lack of measurement, we introduce \methodlong{} (\method{}), which provides agents with measurement telemetry across two turn-level signals: the model's calibrated confidence in its current answer, and whether that answer is entailed by the retrieved evidence. 
Neither signal is sufficient on its own: confidence alone cannot separate a confidently wrong answer from an equally confident (but unverified) correct answer,
and grounding alone cannot separate cases where the agent needs to retrieve more evidence from those where the evidence it has disagrees with the answer. 
~\cref{fig:teaser} shows the resulting failure on one subquestion from the HotpotQA distractor set \citep{yang2018hotpotqa} evaluated on Qwen3-32B \citep{yang2025qwen3}. 
When no telemetry signals are exposed (\emph{left}), the agent cannot distinguish a plausible but weakly supported extraction from a grounded answer, so it stops with the wrong subquestion answer. 
With \method (\emph{right}), low confidence triggers additional retrieval, and low grounding on the retrieved claim triggers refinement rather than commitment. The agent commits only after the candidate answer is supported by the retrieved evidence.

Specifically, our framework pairs a calibrated verbal self-confidence score \citep{wang2025dinco} with a per-claim grounding verifier \citep{tang2024minicheck}, and surfaces both at each turn as \emph{telemetry signals}. 
These calibrated scores are added to the agent's state before each action (one of $\{$\texttt{commit}, \texttt{retrieve}, \texttt{refine}, \texttt{decompose}$\}$). 
Crucially, we let the agent incorporate these scores into its decision-making naturally, without defining specific confidence or grounding thresholds.

We study two complementary ways to incorporate \method{} scores. 
In training-free settings, we simply insert the telemetry signal into the prompt of a frozen agent, thereby steering it toward targeted retrieval or refinement instead of committing to the current action when the available evidence is insufficient.
We demonstrate the flexibility and performance of this approach across 
four QA benchmarks, with the same signal porting to five existing retrieval frameworks (Self-Ask \citep{press2023selfask}, TARG \citep{wang2025targ}, SUGAR \citep{zubkova2025sugar}, Verify-and-Edit \citep{zhao2023verifyandedit}, SeaKR \citep{yao2025seakr}). 
Specifically, it improves F1 scores on four of the five frameworks by up to $+15.4\%$ on TARG and $+7.8\%$ on SeaKR. 
We also show that \method{} can improve trained agents. Specifically, following the reward from Search-R1 \citep{jin2025searchr1}, we train agents backed by open-source models using reinforcement learning (RL). 
We show that, under identical training conditions, agents with states augmented by \method{} achieve higher performance than those without, for both Qwen3-8B and Qwen3-30B-A3B. 
Here, we observe gains of $+5.9\%$ F1 for Qwen3-8B and $+3.3\%$ F1 for Qwen3-30B-A3B on the HotpotQA distractor subset.

\section{Related Work}

Existing approaches expose only a single signal per turn, either confidence or grounding, but not both, so the agent cannot effectively decide which action to take. \emph{Training-free} methods drive action choice from one signal at a time: fixed schedules \citep{trivedi2023ircot}, self-reflection \citep{shinn2023reflexion}, entropy thresholds \citep{jiang2023flare}, complexity routing \citep{jeong2024adaptive}, iterative decomposition \citep{press2023selfask}, or single-signal gates \citep{ding2026cta,zubkova2025sugar,yao2025seakr,vinod2026uncertainty}. \emph{Training-based} methods likewise condition on a single signal, learning the policy end-to-end via outcome reinforcement learning (RL) \citep{jin2025searchr1} or via reflection tokens emitted by the policy itself \citep{asai2024selfrag}. Recent work also exposes verbalized uncertainty to agent loops: Agent-BRACE \citep{singh2026agent} trains a decoupled belief-state model that summarizes history as verbalized, uncertainty-labeled claims about the environment state, which a separately trained policy then conditions on. 
Unlike this learned belief over latent world state, \method's telemetry is training-free and does not encode a belief about the environment, instead exposing two QA-specific external signals about the quality of the agent's current answer.

\paragraph{Cost-aware retrieval gating.}
Prior work gates retrieval using a single confidence score, either fixed before the trajectory or read per turn \citep{jiang2023flare,jeong2024adaptive,wang2025targ,zubkova2025sugar,yao2025seakr,zhao2023verifyandedit}. In contrast, we use calibrated verbal confidence to decide when to retrieve and commit.

\paragraph{Multi-hop agent loops.} ReAct \citep{yao2023react}, IRCoT \citep{trivedi2023ircot}, Self-Ask \citep{press2023selfask}, Self-RAG \citep{asai2024selfrag}, Reflexion \citep{shinn2023reflexion}, and LATS \citep{zhou2024lats} interleave retrieval and reasoning with internal stopping criteria, but do not incorporate external verifier signals. We include calibrated confidence and grounding scores in the agent state before each action so that the agent's decision is conditioned on two orthogonal uncertainty signals.

\section{\texorpdfstring{\methodlong{}}{Method Long Name}}
\label{sec:method}

We incorporate \method{} into a ReAct-style \citep{yao2023react} agent loop. 
A planner first decomposes the question $q$ into a subgoal directed acyclic graph (DAG) $G$, and at each turn, the agent observes the running evidence pool $E$ and chooses one action from a small discrete vocabulary: $\{$\texttt{commit}, \texttt{retrieve}, \texttt{refine}, \texttt{decompose}$\}$. \texttt{Commit} accepts the current answer, stops acting on the current subgoal, and stores that answer along with its supporting facts in memory for the composer to assemble the final answer; \texttt{retrieve} searches for additional evidence passages; \texttt{refine} regenerates the answer from the existing evidence by using a refinement prompt; and \texttt{decompose} gives up on the current subquestion and breaks it into smaller subquestions when retrieval and refinement have not helped. Agents augmented with \method{} receive a telemetry signal in their state representation, informing their subsequent actions.

\paragraph{Verifier Telemetry Scores}
The telemetry is comprised of two channels; each has two values: (1) \textsc{Self-confidence.} The score pair $(\textsc{nvc}, \textsc{sc}) \in [0,1]$ is produced by \dinco \citep{wang2025dinco}: $\textsc{nvc}$ is the normalized verbal confidence from prompting the generator with \emph{``Is this answer correct? Yes/No"}, and $\textsc{sc}$ is the self-consistency rate of the generator. 
(2) \textsc{Grounding family.} A separate verifier model (Bespoke-MiniCheck-7B; \citealp{tang2024minicheck}) decomposes the candidate answer beam $C$ into claims and scores each claim's entailment against the current evidence pool $E$. We extract two values: 
the mean $g_{\text{mean}}$ and the minimum $g_{\min}$ (the worst-grounded claim), with the latter highlighting the weakest link.
This results in 4 total telemetry scalars.

\paragraph{Agent Loop.}
The agent operates over a question $q$, an evidence pool $E$, and a subgoal DAG $G$, produced by the planner. On each turn it picks the next unresolved subgoal $n \in G$, drafts a candidate-answer beam $C$ given $(n, E)$, and chooses one of the four actions. The pair $(\textsc{nvc}, \textsc{sc})$ is computed once when the agent first visits $n$ and cached while the agent remains on $n$; the grounding pair $(g_{\text{mean}}, g_{\min})$ is recomputed every turn over $(C, E)$. The four scalars are provided
in the prompt alongside a brief natural-language framing of what each channel measures; the prompt encodes no hard-coded thresholds, so the calibrated values are inputs the agent reasons over rather than gates baked into the template. 
The loop terminates when $G$ is fully resolved or the per-question turn budget is exhausted, after which a composer pass produces the final answer $y$ from $(q, E)$.

\section{Experiments and Results}
\label{sec:experiments}

\label{sec:setup}

\myparagraph{Datasets.}
We evaluate on four open-domain QA benchmarks: three multi-hop datasets \textbf{HotpotQA-distractor} \citep{yang2018hotpotqa}, \textbf{2WikiMultihopQA} \citep{ho2020twowiki}, and \textbf{MuSiQue} \citep{trivedi2022musique} and the single hop factoid dataset \textbf{WiTQA} \citep{maekawa2024witqa}. We sample $N{=}300$ dev questions per benchmark for evaluation. Retrieval uses BM25 \citep{Robertson1994SomeSE} over each multi-hop distractor pool, and full-Wikipedia BM25 with cross-encoder reranking for WiTQA. For training-time experiments, we train on $1{,}600$ HotpotQA-distractor questions and evaluate on a held-out $N{=}200$ dev slice.

\myparagraph{Metrics.} We report token-level F1 and normalized exact match (EM). To quantify action efficiency, we report mean agent actions per example (turns/ex) and per-action-type counts. Paired conditions are summarized by telemetry minus no-telemetry deltas ($\Delta$F1, $\Delta$turns/ex).

\subsection{Prompt-based Methods}
\label{sec:results-hotpot}

We compare two prompt conditions on a frozen generator running a ReAct-style loop. We have \texttt{+tel}, in which the telemetry scalars and their natural-language framing appear in the per-turn prompt, and \texttt{-tel}, in which both are removed while the remaining prompt is held fixed. 
The planner, retriever, and composer are held fixed across the two. Implementation details are in \cref{sec:imple}.

\paragraph{Setup.} We evaluate the prompt level recipe on both generators across all four benchmarks. The three multi-hop benchmarks (HotpotQA-distractor, 2Wiki, MuSiQue) are evaluated on the agent loop; 
WiTQA uses a single-turn factoid loop with a binary $\{$\texttt{closed-book}, \texttt{retrieve-then-answer}$\}$ decision. 
The pipeline (planner, MiniCheck verifier, telemetry computation, and composer) is fixed; only the action vocabulary and retrieval index change.

\begin{table}[t]
\centering
\footnotesize
\setlength{\tabcolsep}{4pt}
\begin{tabular}{ll ccc c}
\toprule
\multirow{2}{*}{\textbf{Benchmark}} & \multirow{2}{*}{\textbf{Model}} &
\multicolumn{3}{c}{\textbf{F1}} & \textbf{Turns} \\
\cmidrule(lr){3-5}\cmidrule(lr){6-6}
& & \textbf{-tel} & \textbf{+tel} & \textbf{$\Delta\uparrow$} & \textbf{$\Delta$} \\
\midrule
\multirow{2}{*}{HotpotQA}
  & Mistral-24B & 67.9 & 65.2 & \cellcolor[RGB]{249,225,225}$-2.7$ & $-2.61$ \\
  & Qwen3-32B   & 71.1 & 72.5 & \cellcolor[RGB]{229,239,250}$+1.4$ & $-1.26$ \\
\midrule
\multirow{2}{*}{2Wiki}
  & Mistral-24B & 68.0 & 70.8 & \cellcolor[RGB]{208,226,246}$+2.8$ & $-0.60$ \\
  & Qwen3-32B   & 66.0 & 69.7 & \cellcolor[RGB]{191,215,243}$+3.7$ & $-1.87$ \\
\midrule
\multirow{2}{*}{MuSiQue}
  & Mistral-24B & 32.7 & 34.7 & \cellcolor[RGB]{220,233,248}$+2.0$ & $+0.23$ \\
  & Qwen3-32B   & 42.6 & 42.4 & \cellcolor[RGB]{252,242,242}$-0.2$ & $-0.64$ \\
\midrule
\multirow{2}{*}{WiTQA}
  & Mistral-24B & 87.5 & \textbf{89.2} & \cellcolor[RGB]{224,236,249}$+1.7$ & $+0.30$ \\
  & Qwen3-32B   & 82.2 & 86.9 & \cellcolor[RGB]{179,208,241}$+4.7$ & $+0.25$ \\
\bottomrule
\end{tabular}
\caption{Cross-benchmark results; WiTQA adds a CrossEncoder reranker over BM25; $\Delta = \text{with telemetry} \texttt{(+tel)} - \text{without telemetry} \texttt{(-tel)}$ ($\uparrow$ higher is better). \textbf{Turns} reports the $\Delta$ in agent turns per example.}
\label{tab:cross}
\end{table}

\paragraph{Results.} 
\cref{tab:cross} reports per-(benchmark, model) F1 and turns/ex across all four benchmarks. 
On the three multi-hop benchmarks, both generators shrink the action budget under telemetry while F1 holds or improves, with one Mistral-24B trade-off on HotpotQA ($\Delta\text{F1}{=}{-}2.7$, $\Delta$turns/ex${=}{-}2.61$). 
On WiTQA, the same prompt does the opposite: the retrieval rate rises and F1 improves (Qwen3-32B $\Delta\text{F1}{=}{+}4.7$).
This indicates that, when given telemetry signals, the agent is able to adapt to the prevalent failure mode (over-retrieval or parametric over-trust). 

\subsection{Portability}
\label{sec:portability}

\paragraph{Setup.} To test whether the telemetry signal transfers across agent loops without retraining, we evaluate five published adaptive-retrieval frameworks, each with their original prompts and termination conditions. 
Each framework has its own gate for performing retrieval, i.e., a per turn rule that decides whether to retrieve or not. 
We evaluate five frameworks that trigger retrieval based on an uncertainty signal. Per framework, baseline uncertainty and adaptive retrieval details are in \cref{sec:frameworkp}.
We augment each baseline with \method{} by replacing its native retrieval gate with \dinco confidence scores, inserted into the prompt. 
The baseline keeps its original gate; thus, \dinco scores the same candidate answers, and only the uncertainty signal and how it is used change. 
For TARG and SeaKR, we score the closed book question-answer pair using the majority sampled answer. 
Verify-and-Edit scores the current subquestion and majority-sampled answer, while Self-Ask scores the current question and running answer.

\begin{table}[t]
\centering
\footnotesize
\setlength{\tabcolsep}{3pt}
\renewcommand{\arraystretch}{1.18}
\resizebox{\columnwidth}{!}{%
\begin{tabular}{@{}l ccr ccr r@{}}
\toprule
 & \multicolumn{3}{c}{\textbf{EM}} & \multicolumn{3}{c}{\textbf{F1}} & \\
\cmidrule(lr){2-4} \cmidrule(lr){5-7}
\textbf{Framework}
 & \footnotesize -tel & \footnotesize +tel & $\Delta$
 & \footnotesize -tel & \footnotesize +tel & $\Delta$
 & $\Delta$\textbf{Act.} \\
\midrule
V\&E            & 30.0 & 31.0 & $+1.0$ & 38.8 & 41.0 & $+2.2$                  & $+22$ \\
TARG            & 33.0 & 47.0 & $+14.0$ & 45.1 & 60.5 & $\mathbf{+15.4}$        & $+36$ \\
SeaKR & 39.0 & 47.0 & $+8.0$ & 51.0 & 58.7 & $\mathbf{+7.8}$         & $+31$ \\
Self-Ask        & 43.0 & 47.0 & $+4.0$ & 60.3 & 61.7 & $+1.4$                  & $-05$  \\
SUGAR           & 50.0 & 46.0 & $-4.0$ & 61.3 & 60.3 & $-1.0$                  & $\mathbf{-34}$ \\
\bottomrule
\end{tabular}%
}
\caption{Framework portability ($N{=}100$ paired, HotpotQA, Qwen3-32B). The \texttt{+tel} column substitutes \dinco for each framework's gate. $\Delta$\textbf{Act.}\ is the change in total agent actions.}
\label{tab:portability-matrix}
\end{table}

\paragraph{Results.} \cref{tab:portability-matrix} reports per-framework EM, F1, and total change in agent actions. Exposing the telemetry signal at each gate decision boosts F1 on four of the five frameworks; SUGAR's F1 accuracy is essentially unchanged, but its retrieval cost falls by $34$ retrieval calls.

\subsection{Training-time Methods}
\label{sec:grpo-moe}

We next ask whether the agent can better \emph{learn} to choose between the four actions based on outcome feedback.  
Following Search-R1 \citep{jin2025searchr1}, we use GRPO \citep{GRPO} with a verifiable rule based reward to train models for knowledge-intensive QA.

\paragraph{Setup.} We use GRPO to train a LoRA \citep{hu2022lora} adapter (rank $16$) on the per-turn agent via Tinker \cite{tml2026tinker}, with all the other components frozen. We apply this recipe to a dense Qwen3-8B and a Qwen3-30B-A3B Mixture-of-Experts model on a $1{,}600$-question training pool drawn from the HotpotQA distractor dev set and evaluate it on a $200$-question held out set. We also evaluate the same recipe under a telemetry-free prompt, thereby ablating the telemetry signal while keeping everything else identical.  
Hyperparameters are given in \cref{sec:appendix-grpo14b}.
Each rollout is rewarded by the final answer quality minus the action cost. Valid rollouts receive token level F1 against the gold answer, plus a small exact-match bonus, and incur penalties for extra turns and costly actions: retrieval/refinement, decomposition, or malformed actions. Following GRPO, we sample four rollouts per question and reward policies that outperform the group average while using fewer, cheaper actions. See \cref{sec:rewardformula} for the full formulation.

\begin{table}[t]
\centering
\footnotesize
\setlength{\tabcolsep}{3pt}
\renewcommand{\arraystretch}{1.1}
\begin{tabular}{lcccc}
\toprule
 & \multicolumn{2}{c}{\textbf{Qwen3-8B}} & \multicolumn{2}{c}{\textbf{30B-A3B}} \\
\cmidrule(lr){2-3}\cmidrule(lr){4-5}
\textbf{Condition} & EM & F1 & EM & F1 \\
\midrule
Untrained (\texttt{-tel}) & 26.0 & 38.6 & 28.5 & 39.2 \\
GRPO (\texttt{-tel}) & 28.0 & 39.7 & 31.0 & 40.8 \\
GRPO (\texttt{+tel})   & \textbf{35.0} & \textbf{45.6} & \textbf{35.0} & \textbf{44.1} \\
\bottomrule
\end{tabular}
\caption{GRPO-trained policy at 100 steps results on 200 held out HotpotQA-distractor dev questions. Telemetry aware GRPO outperforms telemetry-free GRPO on both model bases.}
\label{tab:grpo8b}
\end{table}

\paragraph{Results.}
\label{sec:grpo-8b}
\cref{tab:grpo8b} reports EM and F1 at step 100. Telemetry-aware GRPO improves Qwen3-8B by $+9.0$\% EM and Qwen3-30B-A3B by $+6.5$\% EM over their untrained baselines. 
The paired ablation isolates telemetry as the active ingredient: telemetry-aware GRPO exceeds telemetry-free GRPO by $+7.0$\% EM / $+5.9$\% F1 on 8B and $+4.0$\% EM / $+3.3$\% F1 on 30B-A3B, while telemetry-free GRPO barely improves.

\subsection{Analysis}
\label{sec:analysis}

\paragraph{Telemetry gains grow with question difficulty.}
\label{sec:analysis-when}
Telemetry routes the agent's action budget toward questions where parametric memory and shallow retrieval fail. \cref{fig:stratified} stratifies per-question $\Delta\text{F1}$ on Qwen3-32B: the HotpotQA gain grows monotonically with hop count, and the WiTQA gain concentrates on tail-popularity subjects while popular subjects are untouched.

\paragraph{Both telemetry signals are essential for better action choices.}
\label{sec:analysis-why}

\cref{tab:channel} groups per-turn telemetry by the Qwen3-32B agent's chosen action: \texttt{commit} correlates with high confidence and high grounding, while \texttt{retrieve} correlates with low confidence and low or absent grounding. This suggests that confidence tracks the agent’s propensity to commit, and grounding tracks commit acceptability.

\paragraph{Prompt-only telemetry depends on model scale.} \label{sec:promptonlywhy} Exposing the signal to a frozen agent helps only at sufficient scale: on a smaller Qwen3-8B base, the prompt-only setting underperforms its no-telemetry baseline, indicating that the agent does not yet act productively on the signal it is shown. 
However, we find that this can be mitigated via training, i.e., in \cref{tab:grpo8b}.

\begin{table}[t]
\centering
\small
\setlength{\tabcolsep}{4pt}
\begin{tabular}{lrcc}
\toprule
\textbf{Action} & \textbf{$n$} & \textbf{mean $\textsc{dinco}$} & \textbf{mean $g_{\min}$} \\
\midrule
\texttt{commit}    & 577 & 0.98 & 0.89 \\
\texttt{retrieve}  & 648 & 0.59 & 0.33 \\
\bottomrule
\end{tabular}
\caption{In HotpotQA telemetry run, Qwen3-32B \texttt{commit}s when confidence and grounding are high, but \texttt{retrieve}s when confidence is low and grounding is weak or absent.}
\label{tab:channel}
\end{table}

\section{Conclusion}

We introduced \method{}, exposing verifier telemetry (turn-level confidence and grounding signals) to QA agents. 
Agents use these signals to decide when to retrieve, refine, decompose, or commit. 
Across prompt-only, framework-substitution, and GRPO training settings, \method{} improves the accuracy-cost tradeoff, reducing redundant actions and encouraging retrieval when appropriate. These gains show that calibrated telemetry provides a simple, framework-agnostic mechanism for better agent control.

\section*{Limitations}
\label{sec:limitations}

By design, the \method{} agent commits as soon as it finds a sufficient answer, which suits single answer multi-hop QA (the setting of all four benchmarks we evaluate) but not answer recall tasks, where the gold answer must contain every relevant item \citep{zhu2024fanoutqa,amouyal2023qampari,min2020ambigqa}.
This failure mode stems from our commit centric action vocabulary rather than from the telemetry signal. Hence, accumulating a complete answer set before terminating is a signal agnostic extension we leave to future work.

\bibliography{custom}

\appendix

\section{Calibration scores of the telemetry signals}

\begin{figure*}[t]
\centering
\includegraphics[width=0.92\textwidth]{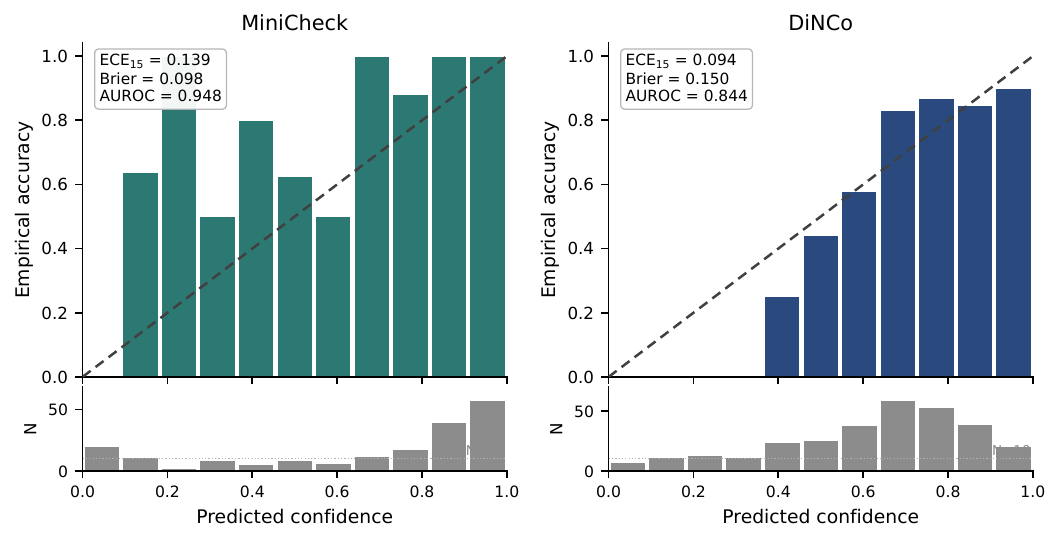}
\caption{Reliability diagrams for the two telemetry signals. \emph{Top:} empirical accuracy per confidence bin, with the dashed line marking perfect calibration. \emph{Bottom:} per-bin sample counts $N$. \textbf{Left} (MiniCheck on AggreFact-RAG): the well populated high-confidence region exists at or above the diagonal, indicating MiniCheck to be confident for correctly grounded answers. \textbf{Right} (\dinco on TriviaQA) bars closely track the diagonal across the well-populated region, yielding strong calibration. }
\label{fig:calibration}
\end{figure*}

For the telemetry signals to be usable in our framework, we make sure that our signals are calibrated by evaluating our confidence and evidence-grounding scoring methods on their task-specific benchmarks. For \dinco confidence signals, we score Qwen3-32B on closed-book questions from a random $N{=}300$ sample (seed=42) of the TriviaQA \citep{triviaqa}  dataset validation split. We choose TriviaQA for this evaluation as its data distribution closely matches the closed-book knowledge regime in which \dinco operates within our framework. For MiniCheck, we draw 300 random (document, claim, label) examples from the LLM-AggreFact dataset, which has aggregated data points from multiple datasets used for the evaluation of grounded factuality of models. We use LLM-AggreFact for evaluation as its distribution matches our intended use case of evaluating the evidence grounding of model claims against retrieved passages. We report ECE-15, Brier, and AUROC scores with 95\% bootstrap confidence intervals over $1{,}000$ resamples. We report our calibration results in \cref{tab:calibration} and reliability plots in \cref{fig:calibration} . \dinco clears the standard $<\!0.10$ ECE-15 bar, while MiniCheck achieves near-perfect discrimination with an AUROC of $0.95$. Through these results, we show that the signals are usable as telemetry in our agent loop.

\begin{table}[h]
\centering
\small
\setlength{\tabcolsep}{3.5pt}
\begin{tabular}{@{}lcc@{}}
\toprule
\textbf{Metric} & \textbf{MiniCheck} & \textbf{DINCO} \\
 & \textsc{aggrefact-rag} & \textsc{triviaqa}/Qwen3-32B \\
\midrule
ECE-15           & $0.139$ $[0.119, 0.189]$ & $\mathbf{0.094}$ $[0.072, 0.143]$ \\
Brier            & $\mathbf{0.098}$ $[0.072, 0.127]$ & $0.150$ $[0.130, 0.173]$ \\
AUROC            & $\mathbf{0.948}$ $[0.913, 0.976]$ & $0.844$ $[0.792, 0.888]$ \\
\bottomrule
\end{tabular}
\caption{Calibration metrics for the two telemetry signals on populations matching \method's regime.
}
\label{tab:calibration}
\end{table}

\section{CIs Across Benchmarks}

For our prompt-only evaluation of \method, we report results on $n=300$ samples evaluated with a single seed value in \cref{tab:cross}. To evaluate whether our setup is robust across varying difficulty levels and question types, we perform a test comparing the 
\texttt{+tel}  and \texttt{-tel} settings on the same question pairs, and report the $95\%$ confidence intervals for each test in \cref{tab:ci}.

For each prompt-only experiment, we use the paired \texttt{+tel} and \texttt{-tel} answers and draw $n_{\text{boot}} = 10{,}000$ paired index resamples from our $N=300$ dev split to compute the confidence intervals. We observe the same F1 improvements and shift in agent action turns in our bootstrap experiment as in the seed run reported in \cref{tab:cross}.

\begin{table*}[t]
\centering
\small
\setlength{\tabcolsep}{5pt}
\begin{tabular}{llcc}
\toprule
\textbf{Benchmark} & \textbf{Model} & $\Delta$\textbf{F1 $\times 100$ (95\% CI)} & $\Delta$\textbf{Turns (95\% CI)} \\
\midrule
\multirow{2}{*}{HotpotQA}
  & Mistral-24B & $-2.7\ [-5.37, -0.06]$           & $\mathbf{-2.61}\ [-3.01, -2.24]$ \\
  & Qwen3-32B   & $+1.4\ [-0.74, +3.55]$           & $-1.26\ [-1.52, -0.99]$ \\
\midrule
\multirow{2}{*}{2Wiki}
  & Mistral-24B & $+2.8\ [-0.63, +6.11]$           & $-0.60\ [-0.89, -0.31]$ \\
  & Qwen3-32B   & $+3.7\ [+1.01, +6.58]$           & $-1.87\ [-2.16, -1.59]$ \\
\midrule
\multirow{2}{*}{MuSiQue}
  & Mistral-24B & $+2.0\ [-2.02, +5.94]$           & $+0.23\ [-0.11, +0.56]$ \\
  & Qwen3-32B   & $-0.2\ [-3.06, +2.59]$           & $-0.64\ [-1.00, -0.26]$ \\
\midrule
\multirow{2}{*}{WiTQA}
  & Mistral-24B & $+1.7\ [-0.48, +4.09]$           & $+0.30\ [+0.25, +0.36]$ \\
  & Qwen3-32B   & $\mathbf{+4.7}\ [+1.20, +8.43]$  & $+0.25\ [+0.20, +0.30]$ \\
\bottomrule
\end{tabular}
\caption{Paired bootstrap 95\% CIs on the eight experiments of ~\cref{tab:cross}.}
\label{tab:ci}
\end{table*}

\section{FLOP analysis of prompt only setups}
\label{sec:appendix-flops}

We analyze the computational overhead in TFLOPs of \method's telemetry addition computed for verbalized confidence metrics using \dinco and grounding confidence scores using MiniCheck. We use torch flop counter to compute the TFLOPs consumed in the \text{tel} and \text{notel} settings by Qwen3-32B and Mistral-24B on two benchmarks. We report our results on 300 samples average TFLOPs in \cref{tab:flops}.

\begin{table}[h]
\centering
\small
\setlength{\tabcolsep}{3.5pt}
\begin{tabular}{@{}llrrrr@{}}
\toprule
\textbf{Bench} & \textbf{Model} & \textbf{$\Delta$F1} & \textbf{$\Delta$turns} & \textbf{$+$DINCO} & \textbf{$\Delta$tot} \\
 & & & & \textbf{TFLOP} & \textbf{\%} \\
\midrule
2Wiki   & Mistral-24B & $+2.8$ & $-0.60$ & 290 & $+222$ \\
2Wiki   & Qwen3-32B   & $+3.7$ & $-1.87$ & 360 & $+159$ \\
MuSiQue & Mistral-24B & $+2.0$ & $+0.23$ & 258 & $+247$ \\
MuSiQue & Qwen3-32B   & $-0.2$ & $-0.64$ & 306 & $+200$ \\
\bottomrule
\end{tabular}
\caption{Compute cost analysis of the prompt-level recipe, results showing deltas computed against \texttt{+tel} and \texttt{-tel}.
F1 deltas reported out of 100.}
\label{tab:flops}
\end{table}

\section{Additional Analysis}

\begin{figure}[t]
\centering
\includegraphics[width=\columnwidth]{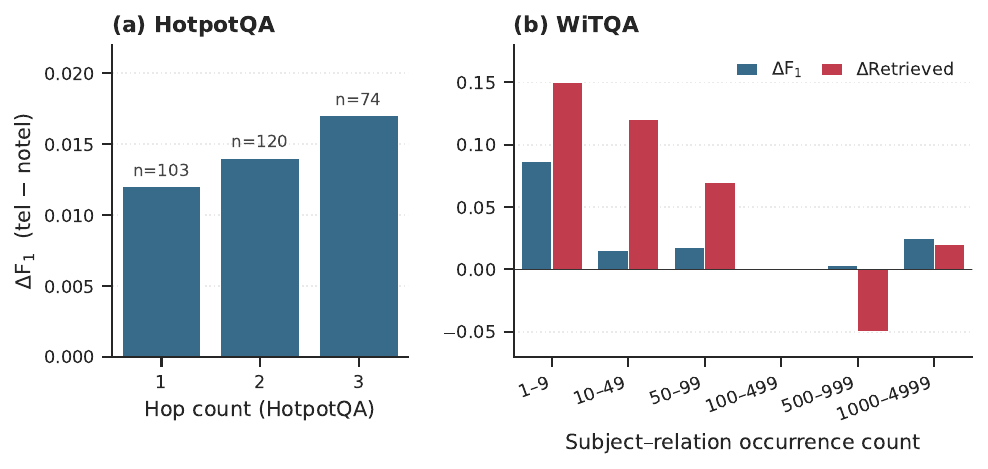}
\caption{ \textbf{Effect of telemetry across question difficulty and entity frequency.} Deltas are computed against \texttt{+tel} and \texttt{-tel} for Qwen3-32B. \textbf{(a)} On HotpotQA, the F1 gain increases with hop count, suggesting that telemetry helps most on harder multi-hop questions. \textbf{(b)} WiTQA gain and retrieval rate increase concentrate on the rarest subject popularity buckets and diminish on popular entities.}
\label{fig:stratified}
\end{figure}

\myparagraph{Telemetry as Action Router.} 
Telemetry, when exposed to harder questions, re-allocates the agent's action budget against the prevailing failure mode rather than uniformly reducing search. On multi-hop benchmarks, adding telemetry shrinks the budget (\cref{tab:cross}); on WiTQA, it expands retrieval for tail entities. (\cref{fig:stratified})

\section{Implementation Details}
\label{sec:imple}
 The prompt-level experiments use two open-weight dense generators, Mistral-Small-24B-Instruct \citep{mistralsmall2025} and Qwen3-32B in non-thinking mode \citep{yang2025qwen3}; the training-time experiments use Qwen3-8B and the Qwen3-30B-A3B Mixture-of-Experts model from the same family \citep{yang2025qwen3}. The grounding verifier is fixed to Bespoke-MiniCheck-7B \citep{tang2024minicheck} across all cells. All generators are served with vLLM \citep{kwon2023vllm} on NVIDIA GH200 nodes. GRPO training runs on the Tinker training service. 

\subsection{GRPO Training Details}
\label{sec:appendix-grpo14b}

\myparagraph{Architecture.} The role-beam runner has four model-driven roles (planner, per-turn role agent, DINCO sampler, composer) plus a hybrid BM25 + bge-base retriever (RRF $k{=}60$, top-$3$). We freeze the planner, DINCO, MiniCheck, and composer on a local vLLM backend and train only the per-turn role agent as a LoRA adapter (rank $16$, lr $1{\times}10^{-4}$, AdamW) hosted on the Tinker training service. Composer and DINCO temperatures match the inference-time settings used elsewhere; the agent rolls out at $T{=}1.0$ with top-$p{=}0.95$. Full GRPO training parameters are reported in \cref{tab:grpo-hparams}.

\myparagraph{GRPO Reward Formulation.}
 \label{sec:rewardformula}
 For a rollout $\tau$ with composer answer $a_{\text{pred}}$ and gold $a_{\text{gold}}$, let $F_1(\tau) = \mathrm{F}_1(a_{\text{pred}}, a_{\text{gold}})$ be the token based F1 reward and $\mathrm{EM}(\tau) = \mathrm{EM}(a_{\text{pred}}, a_{\text{gold}})$ be the EM reward.
 
 A \emph{turn} is one cycle of the agent loop on which the agent emits one action from $\{\texttt{commit}, \texttt{retrieve}, \texttt{refine}, \texttt{decompose}\}$; we let $n_\text{t}$ count turns and $n_\text{r}, n_\text{f}, n_\text{d}$ count the non commit actions within them (commits incur only the base step cost). With $\mathbf{n}(\tau) = (n_\text{t}, n_\text{r}, n_\text{f}, n_\text{d})$ and cost vector $\mathbf{c} = (0.01,\, 0.02,\, 0.02,\, 0.04)$, the reward is
\begin{equation}
\label{eq:grpo-reward}
r_\phi(\tau) =
\begin{cases}
R_{\text{out}}(\tau), & \tau \text{ well-formed}, \\[2pt]
-0.2, & \text{otherwise/malformed},
\end{cases}
\end{equation}
where $R_{\text{out}}(\tau) = F_1(\tau) + 0.1\,\mathrm{EM}(\tau) - \mathbf{c}^{\!\top}\mathbf{n}(\tau)$. For each prompt $q$, we sample $G{=}4$ rollouts $\{\tau_i\}_{i=1}^G \sim \pi_{\theta_\text{old}}(\cdot \mid q)$ and form group-relative advantages $\hat A_i = r_\phi(\tau_i) - \tfrac{1}{G}\sum_{j} r_\phi(\tau_j)$, following GRPO.

\myparagraph{Training Set.} We GRPO train on a $1{,}600$-question pool drawn from the HotpotQA distractor
\emph{dev} split (dev[$0{:}1600$]) rather than the official train split. This
is a difficulty driven choice: the train split is dominated by easy and medium
questions, whereas the released dev split is entirely verified hard questions; \citealp{yang2018hotpotqa}). Our policy
learns \emph{when} to retrieve, refine, decompose, or commit from verifier
telemetry, and this action choice only matters on genuinely hard multi-hop
questions. Training on the mixed train split would spend most updates on
questions the frozen agent already resolves in one or two turns, yielding near zero gradients. We therefore treat this run as a controlled hard regime ablation.

\begin{table}[t]
\centering
\scriptsize
\setlength{\tabcolsep}{4pt}
\begin{tabular}{@{}>{\raggedright\arraybackslash}p{0.34\columnwidth} >{\raggedright\arraybackslash}p{0.56\columnwidth}@{}}
\toprule
\textbf{Hyperparameter} & \textbf{Value} \\
\midrule
\multicolumn{2}{@{}l}{\emph{Models and adapter}} \\
Base models & Qwen3-8B, Qwen3-30B-A3B \\
LoRA rank & $16$ \\
\midrule
\multicolumn{2}{@{}l}{\emph{Optimizer (AdamW)}} \\
Learning rate & $1\times10^{-4}$ (no warmup/schedule) \\
$(\beta_1,\beta_2,\varepsilon)$ & $(0.9,\,0.95,\,10^{-12})$ \\
Weight decay & $0$ \\
Grad accumulation & none ($1$ step/update) \\
\midrule
\multicolumn{2}{@{}l}{\emph{GRPO / PPO}} \\
Group size $G$ & $4$ rollouts / prompt \\
Prompts / step & $8$ \\
Trajectories / step & $32$ \\
PPO epochs / batch & $1$ \\
PPO clip range & $[0.8,\,1.2]$ \\
\midrule
\multicolumn{2}{@{}l}{\emph{Rollout}} \\
Sampling & $T{=}1.0$, top-$p{=}0.95$ \\
Turn budget & $\le 3$ agent turns (first subgoal) \\
\midrule
\multicolumn{2}{@{}l}{\emph{Schedule and selection}} \\
Training steps & $200$,  checkpoints used in eval: 100 \\
Train questions & $1{,}600$ (dev[$0{:}1600$]) \\
Passes over pool & ${\approx}1$ ($200{\times}8$ draws) \\
\bottomrule
\end{tabular}
\caption{GRPO training hyperparameters for the Tinker LoRA. Both bases use identical settings.}
\label{tab:grpo-hparams}
\end{table}

\subsection{Evaluation on Closed Source Models}

\cref{tab:closed-weight-portability} reports results for \texttt{gpt-4o} and \texttt{gpt-4.1} under the \texttt{+tel} and \texttt{-tel} settings, showing that the telemetry signal is not specific to open-weight backbones. We keep the evaluation to 100 random samples from the test set due to budget constraints. \method{} raises F1 on both datasets for both models, with the largest gain on the more retrieval-intensive 2Wiki ($+13.2$ F1 for \texttt{gpt-4o}, $+3.1$ for \texttt{gpt-4.1}) and smaller gains on HotpotQA ($+2.0$ F1 for both). The action turns per example barely move, and the only regression observed is a $-1.0$ EM drop for \texttt{gpt-4.1} on HotpotQA, where overall F1 still improves. This indicates that calibrated telemetry continues to guide action choice in agents even on stronger proprietary models. We limit our closed-source evaluation to older \texttt{gpt} variants because we are unable to evaluate newer closed-source models, which no longer expose top-token log probabilities through their API, a requirement for the pseudo-beam search that \dinco uses to compute its calibrated confidence scores.

\begin{table}[t]
\centering
\footnotesize
\setlength{\tabcolsep}{3pt}
\renewcommand{\arraystretch}{1.18}
\resizebox{\columnwidth}{!}{%
\begin{tabular}{@{}ll ccr ccr r@{}}
\toprule
 & & \multicolumn{3}{c}{\textbf{EM}} & \multicolumn{3}{c}{\textbf{F1}} & \\
\cmidrule(lr){3-5} \cmidrule(lr){6-8}
\textbf{Model} & \textbf{Dataset}
 & \footnotesize \texttt{-tel} & \footnotesize \texttt{+tel} & $\Delta$
 & \footnotesize \texttt{-tel} & \footnotesize \texttt{+tel} & $\Delta$
 & $\Delta$turns/ex \\
\midrule
\multirow{2}{*}{\texttt{gpt-4o}}  & HotpotQA       & 51.0 & 54.0 & $+3.0$  & 63.8 & 65.8 & $+2.0$   & $+0.02$ \\
                                  & 2Wiki  & 32.0 & 42.0 & $+10.0$ & 36.2 & 49.4 & $+13.2$  & $+0.08$ \\
\midrule
\multirow{2}{*}{\texttt{gpt-4.1}} & HotpotQA       & 59.0 & 58.0 & $-1.0$  & 73.8 & 75.8 & $+2.0$   & $+0.17$ \\
                                  & 2Wiki  & 65.0 & 68.0 & $+3.0$  & 75.0 & 78.1 & $+3.1$   & $-0.09$ \\
\bottomrule
\end{tabular}%
}
\caption{Closed weight model results }
\label{tab:closed-weight-portability}
\end{table}

\subsection{Performance with Individual Telemetry}

We perform an ablation study to assess the performance of using just one of the telemetry scores employed in \method at a time. We report our results on the 2Wiki multi-hop dataset and on single-hop WiTQA with $N=300$, using the same seed as in \cref{tab:cross}. We primarily test two setups. In the \dinco-only experiment, \dinco scores are computed for the closed-book setting and passed as telemetry without any post-retrieval scoring. In the MiniCheck-only experiment,  we remove the \dinco closed-book component, allowing the agent to decide what action to take on the closed-book subquestion and then passing grounding scores if retrieval actions are triggered. We report the individual component results alongside the baseline no-telemetry results and the full-telemetry results in \cref{tab:2telab}.

In the multi-hop setting, neither signal in isolation matches the two combined.
\dinco alone is the most detrimental and inflates the number of turns,
suggesting that it leads the agent to overthink and struggle to converge on an
answer. Specifically, we see that without added grounding telemetry in the post-retrieval stage, the agent chose repeated \texttt{decompose} and \texttt{refine} actions until the action budget ran out, thereby leading the composer to give an ungrounded incorrect output at the end.  MiniCheck alone recovers much of this lost performance with fewer turns; yet,
 it still trails the full combination (\method), which attains the highest EM
and F1 while requiring relatively fewer agent turns per example. On the
single-hop WiTQA benchmark, by contrast, all telemetry configurations reach
nearly the same accuracy, yielding only marginal gains over the no-telemetry
baseline. We attribute this to WiTQA being a single-hop factoid benchmark with a
single gold evidence pool, which makes the task substantially easier than the
multi-hop, multi-document setting of 2Wiki. Taken together, these results
indicate that \method delivers effective action routing primarily when a
question demands more complex multi-hop reasoning and retrieval from multiple
sources.

\begin{table}[t]
\centering
\footnotesize
\setlength{\tabcolsep}{4pt}
\begin{tabular}{@{}llrrr@{}}
\toprule
\textbf{Dataset} & \textbf{Config} & \textbf{EM} & \textbf{F1} & \textbf{turns} \\
\midrule
\multirow{4}{*}{2Wiki}
  & \dinco         & 34.2 & 39.2 & 8.44 \\
  & MiniCheck      & 54.0 & 60.4 & 6.04 \\
  & \texttt{-tel}  & 59.3 & 66.0 & 7.36 \\
  & \texttt{+tel}  & \textbf{62.0} & \textbf{69.7} & \textbf{5.49} \\
\midrule
\multirow{4}{*}{WiTQA}
  & \dinco         & 88.0 & 89.2 & 1.42 \\
  & MiniCheck      & \textbf{88.7} & \textbf{89.6} & 1.59 \\
  & \texttt{-tel}  & 86.1 & 87.5 & \textbf{1.19} \\
  & \texttt{+tel}  & 88.4 & 89.2 & 1.49 \\
\bottomrule
\end{tabular}
\caption{Results ($N{=}300$). 2Wiki uses Qwen3-32B; WiTQA uses Mistral-24B. EM and F1 in \%. tel uses both signals.}
\label{tab:2telab}
\end{table}
\subsection{Adaptive Retrieval Frameworks}
\label{sec:frameworkp}

We test the portability of \method on five Adaptive Retrieval Frameworks. TARG  \citep{wang2025targ} reads a prefix logit uncertainty, the mean per token entropy of a short no context draft, and triggers retrieval when it exceeds a threshold $\tau$. TARG is threshold agnostic, so we calibrate  $\tau$ using the quantile rule mentioned in TARG to the target retrieval rate of $\rho=.4$ making $\tau=0.045$. SeaKR~\citep{yao2025seakr} reads an internal-state signal, the regularized log-determinant of the Gram matrix of last-token hidden states across $k=20$ stochastic generations at a mid-network layer, and retrieves when this dispersion exceeds an operating point $\delta=-6$. SUGAR~\citep{zubkova2025sugar} computes a semantic-entropy signal, the entropy over clusters of samples grouped by bidirectional entailment, and splits it with two thresholds into a no-, single-, or multi-step retrieval decision. We use the same quantile rule here to get the thresholds to be $\tau_{lo}=.81$ and $\tau_{hi}=1.0$ . Verify-and-Edit~\citep{zhao2023verifyandedit} samples $N=5$ reasoning chains and verifies-then-retrieves when fewer than $N/2$ of them agree. Self-Ask~\citep{press2023selfask} uses no scalar gate; it iteratively emits follow-up subquestions, retrieving for each until it self-emits a final answer.

\section{Prompts}
\label{sec:appendix-prompts}
\subsection{Role-beam agent system prompt (telemetry)}

The prompt includes the confidence reading \dinco confidence (final\_conf). Following \cref{sec:method}, this is the equal-weight mean of the two self-confidence components,
\begin{equation}
\textsc{final\_conf} \mathrel{:=} \tfrac{1}{2}(\textsc{nvc} + \textsc{sc}) \in [0,1],
\label{eq:final-conf}
\end{equation}
and it is reported as ``mean $\textsc{dinco}$'' in Table~\ref{tab:channel}, averaged over the turns on which the agent took each action. The agent sees both the components $(\textsc{nvc}, \textsc{sc})$ and this summary; \texttt{final\_conf} adds no information beyond them and is shown only so the policy has one calibrated reading without recombining the parts. The grounding family $(g_{\text{mean}}, g_{\min})$ is reported separately.

\label{sec:appendix-prompt-agent}
\begin{lstlisting}[style=prompt]
You are a retrieval policy controller for multi-hop question answering.

You are processing one subquestion at a time as part of answering a larger multi-hop question. After each action you take, the system computes numerical telemetry from confidence and grounding models and shows it to you. You must decide what to do next.

## Available Actions

Return STRICT JSON with exactly one action:

When you commit, you MUST include an "answer" field with your final answer string. Commits without an answer field are invalid.
{"action": "commit", "answer": "final answer string", "analysis": "your analysis of the telemetry", "reason": "why you are committing"}
  Accept the current answer. Use when confidence is high AND the answer is well-grounded in evidence (or when you are confident the answer is correct from parametric knowledge for simple factoid questions).

{"action": "retrieve", "query": "optional search query", "analysis": "your analysis of the telemetry", "reason": "why you need more evidence"}
  Search for more evidence passages. The query defaults to the subquestion text if omitted. Use when confidence is low or grounding is insufficient.

{"action": "refine", "analysis": "your analysis of the telemetry", "reason": "why refinement would help"}
  Re-generate an answer using the same evidence with a refinement prompt. Use when grounding scores suggest the evidence is adequate but the answer or its claims are poorly formulated.

{"action": "decompose", "analysis": "your analysis of the telemetry", "reason": "why this question needs decomposition"}
  Give up on this subquestion and break it into smaller pieces. Use as a last resort when retrieval and refinement have not helped.

## Telemetry Signals

You will receive these numerical signals. Use them to inform your decision:

- DINCO confidence (final_conf): Combined NVC + self-consistency score in [0,1]. Higher means the model is more internally consistent about the answer.
- NVC (normalized verbal confidence): Verbal confidence normalized by how contradictory the alternative beam candidates are.
- SC confidence: Self-consistency -- what fraction of independent samples agree with the main answer.
- MiniCheck grounding (g_mean, g_min): How well the answer's support claims are grounded in retrieved evidence. Only available after retrieval.
- Per-claim support scores: Individual grounding probability for each support claim.

## Decision Principles

1. Think before acting. Explain your reasoning in the "analysis" field before choosing an action.
2. An answer without supporting evidence is risky for non-trivial questions.
3. Diminishing returns. If multiple retrievals haven't helped, decomposition may be the right move.
4. Budget awareness. You have a limited number of turns.

Return STRICT JSON only. No markdown, no extra text outside the JSON object.

\end{lstlisting}

\subsection{Role-beam per-turn prompt}
\label{sec:appendix-prompt-turn}
\begin{lstlisting}[style=prompt]
## Turn {turn}/{max_turns} -- Subquestion: "{subquestion}"

### Original Multi-Hop Question
{original_question}

### Current State
- Current answer: "{answer}"
- Answer source: {source}
- Has dependency memory from earlier subquestions: {has_dependency_memory}
- Evidence passages retrieved so far: {n_passages}

### Closed-Book Confidence Telemetry
{closed_book_confidence_block}

### Grounding Telemetry (MiniCheck)
{grounding_block}

### Action History
{action_history_block}

### Budget
{budget_remaining} turns remaining. Choose your next action wisely.

Return STRICT JSON only: {"action": "...", "analysis": "...", "reason": "..."}
\end{lstlisting}

\subsection{No-telemetry ablation system prompt}
\label{sec:appendix-prompt-notel}
\begin{lstlisting}[style=prompt]
You are a retrieval policy controller for multi-hop question answering.

You are processing one subquestion at a time as part of answering a larger multi-hop question. After each action you take, the system updates the state and you must decide what to do next.

## Available Actions

Return STRICT JSON with exactly one action:

{"action": "commit", "analysis": "your reasoning", "reason": "why you are committing"}
  Accept the current answer. Use when you believe the answer is correct.

{"action": "retrieve", "query": "optional search query", "analysis": "your reasoning", "reason": "why you need more evidence"}
  Search for more evidence passages. The query defaults to the subquestion text if omitted.

{"action": "refine", "analysis": "your reasoning", "reason": "why refinement would help"}
  Re-generate an answer using the same evidence with a refinement prompt.

{"action": "decompose", "analysis": "your reasoning", "reason": "why this question needs decomposition"}
  Give up on this subquestion and break it into smaller pieces. Use as a last resort.

## Decision Principles

1. Think before acting. Explain your reasoning in the "analysis" field before choosing an action.
2. An answer without supporting evidence is risky for non-trivial questions.
3. Diminishing returns. If multiple retrievals haven't helped, decomposition may be the right move.
4. Budget awareness. You have a limited number of turns.

Return STRICT JSON only. No markdown, no extra text outside the JSON object.
\end{lstlisting}

\section{Full Example: agent actions flip upon exposed telemetry}

We show a multi-hop question example evaluated on Qwen3-32B where exposing the telemetry flips the outcome. Without telemetry, the agent commits to the wrong subquestion answer, thereby making the composer emit the incorrect final answer. In the case where telemetry is exposed to the agent, it decides to take the correct actions to retrieve more evidence to converge on the correct answer. The shared hop represents the subquestion for which both agents committed to the right answer. The \texttt{fact} shows the supporting fact that influenced the agent decision.

\begin{lstlisting}[style=trace]
Q (MuSiQue): Who played the girlfriend of the actor who played Marlene McFly
in Back to the Future?
GOLD: Claudia Wells

shared hop: "Who played Marlene McFly?" -> "Michael J. Fox"

================  -tel  No agent telemetry exposed->  "Elisabeth Shue"  [WRONG]  =
sg2  "Who played the girlfriend of Michael J. Fox?"
  t1  obs={answer:"Molly Ringwald"}    act=retrieve
  t2  obs={answer:"Elisabeth Shue"}    act=commit -> "Elisabeth Shue"   # commits first plausible candidate
        fact: "Elisabeth Shue played Jennifer Parker in Back to the Future Part II."
COMPOSER -> "Elisabeth Shue"  [WRONG]

================  +tel  ->  telemetry signals exposed   [CORRECT]  ===========================
sg2  "Who played the girlfriend of Michael J. Fox?"
  t1  obs={answer:"Molly Ringwald", dinco:0.22}   act=retrieve
  t2  obs={answer:"Lea Thompson", g_min:0.35}     act=refine     # low grounding -> keep searching
  t3  obs={answer:"Lea Thompson", g_min:0.35}     act=retrieve
  t4  obs={answer:"Claudia Wells", g_min:0.93}    act=commit -> "Claudia Wells"
        fact: "Jennifer Parker was played by Claudia Wells in Back to the Future."
COMPOSER -> "Claudia Wells"   [CORRECT]
\end{lstlisting}

\label{sec:appendix-tel-example}

\end{document}